\documentclass[11pt]{article}

\usepackage[final]{acl}

\usepackage{times}
\usepackage{latexsym}

\usepackage[T1]{fontenc}

\usepackage[utf8]{inputenc}

\usepackage{microtype}

\usepackage{inconsolata}

\usepackage{graphicx}
\usepackage{amsmath, mathtools, amssymb}
\usepackage{tabularx}
\usepackage{subcaption}
\usepackage{caption}
\usepackage{fancyvrb}
\usepackage{listings}
\usepackage{url}
\usepackage{array}
\usepackage{booktabs}
\usepackage{multirow}
\usepackage{eurosym}
\usepackage{pgfplots}
\usepackage{makecell}
\newcolumntype{C}[1]{>{\centering\arraybackslash}p{#1}}
\newcolumntype{L}[1]{>{\arraybackslash}p{#1}}
\usepackage{xcolor,colortbl}
\usepackage{awesomebox}
\usepackage{floatrow}
\usepackage{hyperref}

\usepackage{pifont}
\usepackage{framed}
\usepackage{capt-of}

\definecolor{OliveGreen}{rgb}{0,0.6,0}
\definecolor{CornellRed}{rgb}{0.7, 0.11, 0.11}
\lstdefinestyle{customJava}{
    language=Java,
    moredelim=[is][\color{OliveGreen}]{\$}{\$}, 
    moredelim=[is][\color{CornellRed}]{@}{@}      
}

\title{Automated Knowledge Component Generation and Interpretable
Knowledge Tracing in Coding Problems}

\author{
 \textbf{Zhangqi Duan\textsuperscript{1}},
 \textbf{Nigel Fernandez\textsuperscript{1}},
 \textbf{Arun Balajiee Lekshmi Narayanan\textsuperscript{2}},
 \\
 \textbf{Mohammad Hassany\textsuperscript{2}},
 \textbf{Rafaella Sampaio de Alencar\textsuperscript{2}},
 \textbf{Peter Brusilovsky\textsuperscript{2}},
 \\
 \textbf{Bita Akram\textsuperscript{3}},
 \textbf{Andrew Lan\textsuperscript{1}},
\\
 University of Massachusetts Amherst \textsuperscript{1},
 University of Pittsburgh \textsuperscript{2},
 \\
 North Carolina State University \textsuperscript{3}
\\
\href{mailto: zduan@cs.umass.edu}{zduan@cs.umass.edu}
}

\begin{document}
\maketitle
\begin{abstract}
  Knowledge components (KCs) are key to assessing student knowledge levels on fine-grained skills and driving personalization and feedback. However, crafting KCs and tagging them for problems, traditionally performed by human domain experts, is highly labor-intensive. Prior work has studied automated KC generation only for multiple-choice questions but not open-ended ones. We bridge this gap and present an automated, large language model (LLM)-based pipeline for KC generation and tagging for open-ended programming problems. We also develop an LLM-based knowledge tracing (KT) framework to leverage these LLM-generated KCs. We conduct extensive quantitative and qualitative evaluations on two real-world student code submission datasets. Results show that our KT method outperforms existing ones and LLM-generated KCs outperform human-written KCs on future student response prediction. We also investigate how these KCs enable us to analyze student learning curves and conduct human evaluation with course instructors to further verify the quality of KC-problem tagging. 
\end{abstract}

\section{Introduction} 
In student modeling, an important task is to map problems (or items) to specific skills or concepts, referred to as knowledge components (KCs). KCs provide an invaluable resource to model student learning~\cite{bier2014approach}, estimating their mastery levels~\cite{kt} on fine-grained units of knowledge. Accurately estimating student mastery levels on KCs helps enable both 1) teacher feedback, by showing this information in teacher dashboards,
and 2) adaptive and personalized learning in online learning platforms or intelligent tutoring systems~\cite{huang2020general}, by tailoring instructions and content sequencing according to student knowledge levels. Identifying fine-grained KCs students struggle~\cite{rivers2016learning} with also enables content designers to develop targeted instructional content and practice problems for students. Therefore, KC mastery estimation is fundamental to student modeling for personalization and recommendation.

KCs are typically crafted by human domain experts, who also tag problems with KCs that students need to master to solve the problem correctly. This process can be highly labor-intensive and prone to bias and errors. There exist solutions to automate parts of this process, usually employing classification algorithms~\cite{pardos2017imputing} to tag KCs to problems, which rely on having a predefined set of KCs. Recent advances in Large Language Models (LLMs) have shown potential in developing automated approaches for KC \emph{identification} in addition to tagging, in domains such as math~\cite{ozyurt2024automated} and science~\cite{moore2024automated}. Automatically generating KCs is challenging since KCs need to satisfy various criteria, including being relevant to problems, being specific enough to provide teacher and student support, and being generalizable across settings. Another important aspect is that they need to satisfy cognitive science principles, i.e., student error rates on a KC should decrease as they attempt it more times, according to the power law of practice \cite{snoddy1926learning}. 

\begin{table*}[t]
\centering
\renewcommand{\arraystretch}{1.1}
\scalebox{0.92}{
\begin{tabular}{p{0.45\linewidth}|p{0.28\linewidth}|p{0.2\linewidth}}
\toprule
\multicolumn{3}{p{1.05\linewidth}}{Problem: The number 6 is a truly great number. Given two int values, a and b, return true if either one is 6. Or if their sum or difference is 6. Note: the function Math.abs(num) computes the absolute value of a number.}\\
\midrule
Representative Solution Code & Generated KCs & Human-written KCs\\
\midrule
\multirow{-1.8}{*}{\begin{lstlisting}
public boolean love6(int a, int b){
    if (a == 6 || b == 6){
        return true;
    }
    else if ((a + b) == 6 || Math.abs(a - b) == 6){
        return true;
    }
    else{
        return false;
    }
}
\end{lstlisting}
} & If and else if statement & If/Else \\
& Basic arithmetic operations & Math ($+-*/$) \\
& Logical operators & LogicAndNotOr \\
& Numerical comparisons & LogicCompare \\
& Absolute value computation &  \\
\bottomrule
\end{tabular}}
\vspace{-.3cm}
\caption{Example programming problem from the CodeWorkout dataset with a sample student solution code, comparing KCs generated by our KCGen-KT framework to human-written KCs.}
\label{tab:qualitative_example}
\end{table*}

One notable limitation in existing automated KC generation methods is that they are developed only for multiple-choice questions, which have a rigid, fixed structure. On the contrary, open-ended problems that require students to submit open-ended responses, is more challenging for automated KC generation: there is significant diversity among student responses; students may be incorrect in many different ways and there may even exist numerous different correct solution approaches, each covering a different set of KCs. Take open-ended programming problems that are common in computer science education as an example: students can use different programming concepts to implement the same desired functionality (see Table~\ref{tab:mot_example} in Appendix~\ref{example: motivation} for a concrete example). Therefore, despite approaches leveraging the syntactic structure of code for KC generation \cite{umap,hosseini2013javaparser,Pavlik_Koedinger}, automated KC generation for such open-ended problems remains under-studied.


\subsection{Contributions}

In this paper, we explore using LLMs to automatically generate KCs for open-ended programming problems. 
Our contributions are as follows:

First, we develop an automated, LLM-based pipeline for KC generation and tagging. We first select a diverse set of representative student code submissions to each problem and then prompt GPT-4o to identify KCs that are required to solve the problem. Then, to aggregate KCs across problems and de-duplicate similar ones, we cluster KCs on semantic similarity, followed by summarizing each cluster into a KC description. Finally, we automatically tag problems with KCs according to the clustering results. Table~\ref{tab:qualitative_example} shows an example problem and associated KCs. 

Second, we develop an LLM-based KT method, KCGen-KT,\footnote{Our code can be found at: \url{https://github.com/umass-ml4ed/kcgen-kt}.} to leverage the textual descriptions of the KCs for future student response prediction. Our method explicitly captures student mastery levels on each KC, enabling interpretability. We introduce a soft-token conversion mechanism to inject the mastery level to the LLM input space for gradient flow through models.

Third, we conduct extensive quantitative and qualitative evaluations on two datasets that contain real-world student code submissions to open-ended programming problems. Results show that KCGen-KT outperforms existing KT methods and human-written KCs on predicting future student performance. We also show that LLM-generated KCs are comparable to human-written KCs in terms of learning curves under a commonly used student model. We also conduct a human evaluation to show that the KC tagging accuracy of our pipeline is reasonably accurate to human instructors.

We acknowledge up front that we have no choice but to narrowly focus on programming problems, due to data availability: to study both KC generation and downstream student modeling, we need to not only i) study a large collection of open-ended student responses but also ii) track the same student across multiple problem attempts over time. The only large-scale, publicly available datasets all contain student responses to programming problems in computer science education. Due to space limitations, please see Section~\ref{sec:rw} in the Appendix for a more detailed review of related work.

\begin{figure*}[t]
\centering
\includegraphics[width=\textwidth]{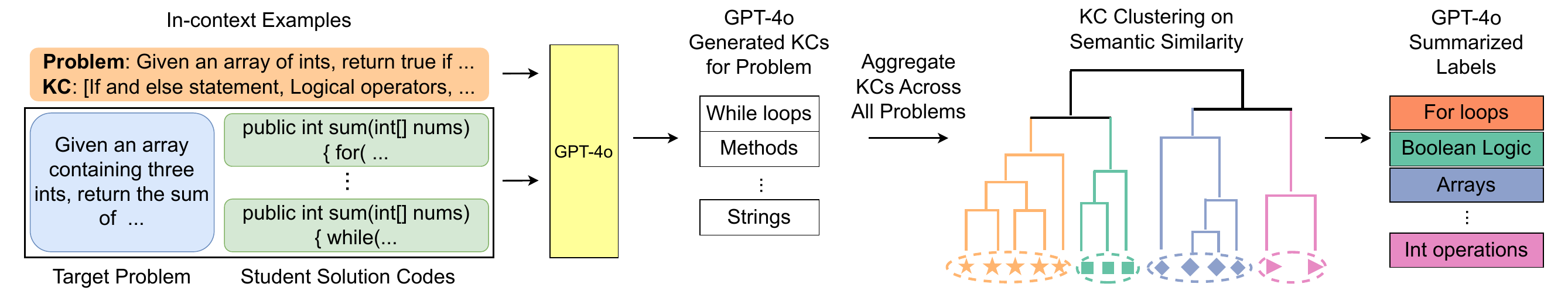}
\vspace{-.5cm}
\caption{Illustration of our three-step automated KC generation and tagging pipeline.} 
\label{fig:kc_gen}
\end{figure*}

\section{Methodology}
We now detail our automated LLM-based approach to generate KCs for programming problems, and then introduce KCGen-KT, a strong KT method leveraging the semantics of the generated KCs to improve student performance prediction.

\subsection{Automated KC Generation} 
For KC generation, we use GPT-4o \citep{gpt-4o}, an advanced proprietary LLM with strong reasoning and programming abilities. Illustrated in Figure~\ref{fig:kc_gen}, we generate KCs for a programming problem following three key steps: 1) generating KCs associated with each problem and their descriptions separately through few-shot prompting, 2) clustering KCs across all problems, and 3) summarizing each cluster to obtain a finalized description of each KC. All steps are performed independently for each dataset to maintain domain consistency. We detail these steps below. 

\noindent \textbf{Initial KC Generation} For each programming problem, we prompt GPT-4o using chain-of-thought reasoning to generate a list of KCs that capture the underlying skills necessary to solve the problem. We include a few carefully constructed in-context examples in our prompt as few-shot demonstrations, prompting GPT-4o to convert human-written topic tags, such as ``If-else'', from each dataset into more fine-granular natural language descriptions. We instruct the model to first explain the relevance of each KC, and then generate a clear textual description. To help the LLM better understand what is required to solve a problem, we include correct student submissions as examples. Since programming problems can often be solved in multiple valid ways, we include diverse examples to ensure comprehensive coverage of relevant KCs. Therefore, we apply a clustering algorithm to the CodeBERT \citep{feng-etal-2020-codebert} embeddings of all correct student submissions and sample one per cluster, with the number of clusters controlling the diversity of examples. Empirically, we find that this approach yields fine-grained, function-level KC descriptions. See Appendix \ref{sec: prompt} for the exact prompt used for all steps in our KC generation pipeline. \\

\noindent \textbf{Clustering KCs and Controlling Abstraction Level} The KCs generated for each problem are initially fine-grained, often describing specific function-level skills or concepts. To control the abstraction level and obtain more generalizable KC descriptions, we first compute the Sentence-BERT \citep{reimers-2019-sentence-bert} embedding of the textual description of each KC, then apply Hierarchical Agglomerative Clustering (HAC) using cosine similarity as the distance function. By adjusting the number of clusters, we can flexibly merge semantically similar KCs into broader categories, effectively controlling the abstraction level of our KC descriptions. This clustering process enables us to move from detailed skill-level descriptions to higher-level conceptual groupings, aligning the KCs with different pedagogical or analytical goals depending on the downstream application. \\

\noindent \textbf{Labeling KC Clusters} Finally, we label each KC cluster by prompting GPT-4o to generate a single, informative name that represents the cluster. We use a chain-of-thought prompt that guides the model to first reason whether any KC in the cluster can represent the entire group. If such a KC exists, it is selected as the cluster label; otherwise, the model is instructed to synthesize a concise description that captures the shared meaning of KCs in the cluster. This process yields the final set of generated KCs across problems at the desired level of abstraction. As a final step, we obtain problem-KC mappings, i.e., a Q-matrix \cite{barnes2005q}, by mapping each initially generated KC for each problem to its corresponding summarized cluster label.

\subsection{Improving Knowledge Tracing via LLM-Generated KCs} \label{2.2}
We now detail KCGen-KT, a novel LLM-based KT method that leverages KC semantics and explicitly models student mastery levels on each KC, for improved KT performance. \\

\noindent \textbf{KT Problem Formulation} For open-ended programming problems, we define each student response to a problem as $x_t \coloneq (p_t, \{w^i_t\}, c_t, a_t)$, where $p_t$ is the textual statement of the problem, $\{w^i_t\}$ are the KCs associated with the problem, $c_t$ is the student code submission, and $a_t$ is the correctness of the submission; in most existing KT methods, $a_t$ is treated as binary-valued (correct/incorrect). 
Therefore, our goal is to estimate a student's mastery level of each KC given their past responses, $x_0, \ldots, x_t$, and use this estimate to predict both 1) the overall binary-valued correctness $a_{t+1} \in \{0,1\}$ and 2) the open-ended code $c_{t+1}$ submitted by the student on their next attempted problem $p_{t+1}$. Following previous work~\cite{codedkt}, $a_t=1$ if the student-submitted code passes all test cases associated with the problem, and $a_t=0$ otherwise. \\

\begin{figure*}
\includegraphics[width=\textwidth]{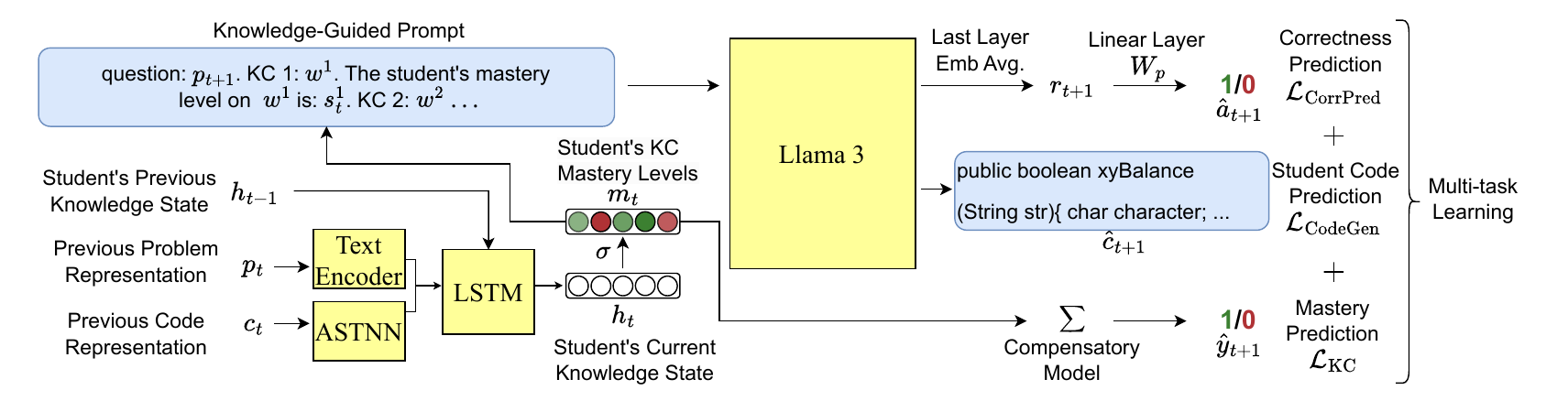}
\vspace{-.7cm}
\caption{Overview of our KCGen-KT's model with the Llama 3 LLM as the backbone. KCGen-KT leverages KC semantics, tracking student mastery levels on each KC, to predict both correctness and the student code submission.} 
\label{fig:model}
\end{figure*}

\noindent \textbf{KCGen-KT}
KCGen-KT leverages the KCs associated with a problem in two ways: 1) by improving the problem representation using the semantic information of KCs, and 2) by improving the student representation by building an interpretable student profile that models student mastery levels on KCs. 

Following TIKTOC~\cite{tiktoc}, we use an open-source LLM, Llama 3~\cite{llama3}, as the backbone to predict both the overall correctness and actual open-ended student code in a token-by-token manner, in a multi-task learning approach. KCGen-KT differs from OKT~\cite{okt} by leveraging the content of the KCs, and from Code-DKT~\cite{codedkt} by using text embedding methods to embed the textual problem statement. \\

\noindent \textbf{Student Knowledge on KCs} For each student, at each timestep $t$, KCGen-KT updates the student's $512$-dimensional knowledge state vector $h \in \mathcal{R}^{512}$, through a long short-term memory (LSTM)~\cite{lstm} network as in DKT~\cite{dkt}, given by $h_{t} = \text{LSTM}(h_{t-1}, p_t, c_t)$.
This knowledge state $h_t$ is compressed into a $k$-dimensional mastery vector $m_t \in [0,1]^{k}$, where $k$ is the total number of KCs, through a linear layer with weights $W_{m}$ and bias $b_{m}$, followed by a sigmoid function to map the values of $m_t$ to be in the range of $[0,1]$, given by $m_t = \sigma(W_{m}h_t + b_m)$. Each dimension $j$ of $m_t$ denotes a student's mastery level on the $j$th unique KC, with larger values denoting higher mastery. \\

\noindent \textbf{Soft KC Mastery Token Conversion} To use LLMs to predict the student response to the next problem, we need to connect student KC knowledge with the textual input space of LLMs. Therefore, following previous work~\cite{fernandez-etal-2024-divert,liu2023boltforsofttokens}, we transform KC mastery levels into \emph{soft} text tokens, i.e.,
\begin{align*}
    s_t^j = m_t^j\cdot \text{emb}^{\text{true}} + (1-m_t^j)\cdot \text{emb}^\text{{false}},
\end{align*}
where $\text{emb}^{\text{true}}$ and $\text{emb}^\text{{false}}$ are the embeddings of the text tokens ``true'', and ``false'', respectively. In other words, we use student KC mastery levels $m_t^j$ to combine two hard, discrete text tokens (``true'' and ``false'') into a differentiable soft token $s_t^j$, to enable the flow of gradients during training. We pass this student knowledge information using the input format of
\texttt{KC $1$: <$w^1$>. The student's mastery level on <$w^1$> is: $s_t^{1}$. $\ldots$} into the LLM for prediction. \\

\noindent \textbf{Knowledge-Guided Response Prediction} We construct our LLM prompt for the next response prediction by including both 1) the textual statement of the next problem and 2) student mastery levels on the KCs associated with the problem, as
\texttt{question: {$p_t$}. <KCs with student mastery levels>}, as shown in Figure~\ref{fig:model}.

To predict the binary-valued correctness of the next student response, we average the hidden states of the last layer of Llama 3 that correspond to only the input (knowledge-guided prompt) to obtain a representation $r$, transformed for correctness prediction using a linear transformation matrix $W_p$ and a sigmoid function, given by $\hat{a}_{t+1} = \sigma(W_p \cdot r)$. We minimize the binary cross-entropy (BCE) loss, which for one response is given by:
\begin{equation*} 
    \mathcal{L}_{\text{CorrPred}} \!=\! a_{t+1} \cdot \log \hat{a}_{t+1} + (1-a_{t+1})\cdot \log (1-\hat{a}_{t+1}).
\end{equation*}
To predict student code, we feed the knowledge-guided prompt into Llama 3 to generate the predicted code $\hat{c}$ token by token. We minimize
\begin{equation*} 
    \mathcal{L}_{\text{CodeGen}} = \textstyle\sum_{n=1}^N \! -\log P_{\theta} \! \left( \hat{c}^n\ \middle \vert\ p,j,\{\hat{c}^{n'}\}_{n'=1}^{n-1}\right)\!,
\end{equation*}
where N is the number of tokens in the student code. $\theta$ denotes the set of learnable parameters, which includes the KT model, the linear layer with weights $W_{m}$ and bias $b_{m}$ for student mastery levels, and the parameters of the finetuned Llama 3.\\ 

\noindent \textbf{Promoting Interpretability} To promote interpretability of student KC knowledge parameters, we use a compensatory model~\cite{maier2021challenges} and take the average of individual student KC mastery levels to obtain an overall mastery level \\
\mbox{$\hat{y}_{t+1} = \frac{1}{\sum_{k=1}^K \mathbb{I}(w_k)}\sum_{k=1}^{K} m_t^k\cdot \mathbb{I}(w_k)$}, where the indicator function $\mathbb{I}(w_k)$ is $1$ if the KC $w_k$ is associated with the problem, and $0$ otherwise. Empirically, we found that averaging over KC masteries performed better than taking the product over them, consistent with findings in prior work \citep{maier2021challenges}. We then minimize the BCE loss between the overall KC mastery and its binary correctness,
\begin{equation*} 
    \mathcal{L}_{\text{KC}} = a_{t+1} \cdot \log \hat{y}_{t+1} + (1-a_{t+1})\cdot \log (1-\hat{y}_{t+1}).
\end{equation*}
This loss regularizes the model to be monotonic, i.e., high knowledge on KCs corresponds to a high probability of a correct response, thus promoting the interpretability of $m_t^j$.  \\

\noindent \textbf{Multi-task Learning Objective} Following previous work~\cite{tiktoc} showing multiple objectives in KT are mutually beneficial to each other, our final multi-task training objective minimizes a combination of all three losses together, with a balancing parameter $\lambda \in [0,1]$ controlling the importance of the losses, as
\begin{equation*}
    \mathcal{L}_{\text{KCGen-KT}} = \lambda (\mathcal{L}_{\text{CodeGen}} + \mathcal{L}_{\text{CorrPred}}) + (1-\lambda) \mathcal{L}_{\text{KC}},
\end{equation*}
where losses are averaged over code submissions by all students to all problems.

\begin{table*}[t]
\centering
\vspace{-0.3cm}
\scalebox{.9}{
\begin{tabular}{p{0.12\linewidth}p{0.33\linewidth}p{0.12\linewidth}p{0.12\linewidth}p{0.12\linewidth}|p{0.13\linewidth}}
\toprule
\multirow{3}{*}{Dataset} & \multirow{3}{*}{Model} & \multicolumn{3}{c}{KT Correctness Pred.} &  \multicolumn{1}{c}{Code Pred.}\\
\cmidrule{3-6}
& & AUC $\uparrow$ & F1 Score $\uparrow$ & Accuracy $\uparrow$ & CodeBLEU $\uparrow$\\
\midrule
\multirow{6}{*}{\shortstack{CodeWorkout \\ (Java)}}
 & Random &  0.499 & 0.368 & 0.506 & $-$ \\
 & Majority &  0.500 & 0.644 & 0.526 & $-$ \\
 & Code-DKT~\cite{codedkt} &  $0.766_{\pm 0.018} $ & $0.672_{\pm 0.033} $ & $0.724_{\pm 0.010} $ & $-$ \\
 & TIKTOC*~\cite{tiktoc} & $0.788_{\pm 0.013}$ & $0.666_{\pm 0.030}$ & $0.726_{\pm 0.013}$ & $0.507_{\pm 0.015}$\\
 & KCGen-KT (Human-written KCs) & $\underline{0.797}_{\pm 0.016}$ & $\underline{0.706}_{\pm 0.026}$ & $\underline{0.727}_{\pm 0.020}$ & $\underline{0.557}_{\pm 0.028}$\\
 & KCGen-KT (Generated KCs) & $\textbf{0.816}\textsuperscript{$\bullet$}_{\pm 0.012}$ & $\textbf{0.727}\textsuperscript{$\bullet$}_{\pm 0.027}$ & $\textbf{0.746}\textsuperscript{$\bullet$}_{\pm 0.012}$ & $\textbf{0.580}\textsuperscript{$\bullet$}_{\pm 0.018}$\\
\midrule
\multirow{6}{*}{\shortstack{FalconCode \\ (Python)}}
 & Random & 0.502 & 0.435 & 0.498 & $-$ \\
 & Majority & 0.500 & 0.547 & 0.603 & $-$ \\
 & Code-DKT~\cite{codedkt} & $0.709_{\pm 0.013}$ & $0.552_{\pm 0.021}$ & $0.617_{\pm 0.019}$ & $-$ \\
 & TIKTOC*~\cite{tiktoc} & $0.728_{\pm 0.011}$ & $0.585_{\pm 0.019}$ & $0.633_{\pm 0.006}$ & $0.427_{\pm 0.010}$ \\
 & KCGen-KT (Human-written KCs) & $\underline{0.752}_{\pm 0.005}$ & $\underline{0.599}_{\pm 0.013}$ & $\underline{0.700}_{\pm 0.008}$ & $\underline{0.473}_{\pm 0.007}$ \\
 & KCGen-KT (Generated KCs) & $\textbf{0.771}\textsuperscript{$\bullet$}_{\pm 0.011}$ & $\textbf{0.645}\textsuperscript{$\bullet$}_{\pm 0.010}$ & $\textbf{0.712}\textsuperscript{$\bullet$}_{\pm 0.006}$ & $\textbf{0.498}\textsuperscript{$\bullet$}_{\pm 0.009}$ \\
\bottomrule
\end{tabular}}
\vspace{-0.3cm}
\caption{Comparing KCGen-KT against baselines on KT performance across all metrics. KCGen-KT, especially with LLM-generated KCs, outperforms other KT methods. Best performance is in \textbf{bold} and second best is \underline{underlined}. \textsuperscript{$\bullet$} denotes statistically significant improvement over baselines ($p<0.05$).}
\label{tab:kt_results}
\end{table*}

\section{Experimental Evaluation}
We now detail our quantitative experimental settings for evaluating KCGen-KT in predicting future student code and correctness. \\

\noindent \textbf{Dataset Details} The CodeWorkout~\cite{codeworkout} dataset was first used in the Second CSEDM Data Challenge~\cite{csedm} and contains actual open-ended code submissions from students, collected from an introductory \textit{Java} programming course, together with problem textual statements and human-written KC tags (estimated programming concepts) from a fixed set of $18$ KCs on each problem. In total, there are $246$ students attempting $50$ problems covering various concepts. Following prior work \cite{codedkt} and standard KT setup, we analyze students' first submissions to each problem, leading to a total of $10{,}834$ submissions.
The FalconCode ~\cite{10.1145/3545945.3569822} dataset similarly consists of open-ended student submissions in an introductory \textit{Python} programming course. The dataset provides problem statements and instructor-labeled KC tags from a predefined set of $20$ KCs. There are $3{,}267$ students attempting $157$ problems in both skill-based and lab-style formats. Similar to CodeWorkout, we analyze students' first submission to each problem, resulting in a total of $28{,}617$ code submissions. \\

\noindent \textbf{Metrics} Following~\cite{codedkt}, we use \textbf{AUC}, \textbf{accuracy}, and \textbf{F1} score to evaluate the correctness prediction performance. For the student code prediction task, following~\cite{okt}, we measure the similarity between generated student code and ground-truth student code using \textbf{CodeBLEU}~\cite{codebleu}, a variant of the classic text similarity metric BLEU~\cite{papineni2002bleu}. This metric is customized for code and measures both syntactic and semantic similarity. \\

\noindent \textbf{Baselines} In terms of KCs, we compare our generated KCs against human-written KCs that are available in both datasets. We test a version of KCGen-KT by replacing our LLM-generated KCs with human-written KCs and keeping the KT method unchanged, which we refer to as \textbf{KCGen-KT (Human-written KCs)}. 
In terms of KT methods, we adapt Test case-Informed Knowledge Tracing for Open-ended Coding (TIKTOC)~\cite{tiktoc}, a recent, strong KT method for programming, as the main baseline. TIKTOC also uses Llama 3 as the backbone and a multi-task learning setup to jointly predict the exact code token-by-token and whether it passes each test case. We slightly modify it for our KT task, replacing test case prediction with overall code correctness prediction, by reducing the dimension of the prediction head from the number of test cases to one, for overall correctness prediction only. We refer to the resulting method as \textbf{TIKTOC*}. We also use \textbf{Code-DKT}, a popular KT method for programming that leverages the content of student code, to predict the overall correctness of student code submissions. As a sanity check, to estimate a lower bound of performance on our KT task thereby providing a sense of task difficulty, we include two simple baselines: \textbf{Random}, which simply predicts the overall binary-valued correctness of a student code randomly with equal probability, and \textbf{Majority}, which simply predicts the majority correctness label (incorrect) among students for each problem.  \\

\noindent \textbf{Experimental Setup} For the KT method component of KCGen-KT as well as for all KT baselines, to ensure a fair comparison, we use the instruction-tuned version of Llama 3~\cite{llama3} with $8$B parameters as the base LLM and a frozen ASTNN~\cite{astnn} as the code embedding model. See Appendix \ref{sec: experiment} for detailed parameter settings. We repeat our experiments across 5 random train-validation-test data splits.

\section{Results, Analysis, and Discussion}
We now discuss our quantitative evaluation results and qualitatively analyze the estimated student KC mastery and predicted code. We also analyze the learning curves, conduct an ablation study, and investigate the patterns of KCs across different levels of abstraction. 

\subsection{Quantitative Evaluation}
\begin{table}[ht]
\centering
\vspace{-0.3cm}
\small
\scalebox{.78}{
\begin{tabular}{l l cccc}
\hline
Dataset & Clusters (Lvl) & AUC & F1 & Acc & CodeBLEU\\
\hline
 & 100 (Low)   & \underline{0.815} & \underline{0.726} & \underline{0.737} & 0.572\\
  & 75 (Low)  & 0.810 & 0.719 & 0.735 & \underline{0.575} \\
CodeWorkout & 50 (Medium) & \textbf{0.816}    & \textbf{0.727}    & \textbf{0.746}    & \textbf{0.580} \\
 & 20 (High)   & 0.804 & 0.698 & 0.710 & 0.561\\
 & 10 (High)  & 0.794             & 0.683             & 0.708             & 0.557 \\ 
\hline
 & 150 (Low)  & 0.763 & 0.628 & 0.675 & 0.494 \\
 & 120 (Low) & \underline{0.763}  & \underline{0.651} & \textbf{0.706} & 0.483 \\
FalconCode & 90 (Medium) &  0.762  & 0.648  &  0.674  & \underline{0.497} \\
 & 60 (Medium)   &   \textbf{0.767}  &   \textbf{0.653}   &  \underline{0.701}   & \textbf{0.503} \\ 
  & 20 (High)  &  0.750   &    0.623     &   0.663   & 0.473 \\ 
\hline
\end{tabular}
}
\vspace{-0.3cm}
\caption{Comparing different KC abstraction levels across datasets. SD omitted for space.}
\label{tab:abstraction}
\end{table}

\noindent \textbf{KCGen-KT outperforms baselines} Table~\ref{tab:kt_results} shows the average performance (and standard deviation) on our two KT tasks: binary correctness prediction and student code generation for all methods on both datasets. For KCGen-KT, we report results using the best-performing configuration for each dataset, which generates $50$ KCs after clustering for CodeWorkout and $60$ KCs after clustering for FalconCode, based on $5$ student submissions per problem. Across both datasets, we see that the Random and Majority baselines perform poorly, which suggests that the correctness prediction KT task is inherently difficult. 
Our proposed framework, KCGen-KT with either human-written or generated KCs, outperforms other strong KT methods that do not use KCs, including TITKOC* and Code-DKT. This observation suggests that for KT methods that use LLMs as the backbone, leveraging the semantic information in KC descriptions improves KT performance. More importantly, KCGen-KT with our generated KCs outperforms human-written KCs by a consistent margin on both datasets and across both tasks, with statistical significance ($p < 0.05$). This observation shows that high-quality KC descriptions and accurate tagging are key to improving downstream KT performance. The performance gap is more evident in code prediction, which shows that semantically informative KCs, as evident from Table~\ref{tab:qualitative_example}, are especially important to LLMs in generative tasks.  \\

\noindent \textbf{Less Fine-grained KCs hurt performance} To investigate the impact of KC granularity on model performance, we experiment with multiple abstraction levels across two datasets. For the CodeWorkout dataset, our pipeline first generates $103$ unique KCs, then applies the clustering algorithm with the number of clusters set to $100$, $75$, $50$, $20$, and $10$. For the FalconCode dataset, we begin with $161$ unique KCs and then cluster them into $150$, $120$, $90$, $60$, and $20$ clusters. As the number of clusters decreases, the resulting KC sets become increasingly abstract, forming a hierarchy of representations. We evaluate KCGen-KT's performance under all abstraction levels. Table~\ref{tab:abstraction} shows that the highest abstraction levels consistently lead to the lowest performance across all metrics and datasets. In contrast, medium and low abstractions achieve comparable performance, which justifies our choice of using medium-level KCs in the main experiments. These results suggest that overly abstract KCs may fail to pinpoint necessary skills in a problem, underscoring the importance of having sufficient granularity in KCs for student modeling tasks.


\begin{table}[h]
\centering
\vspace{-0.1cm}
\small
\scalebox{1}{
\begin{tabular*}{\linewidth}{lcccc}
\hline
Model & AUC & F1 & Acc & CodeBLEU\\
\hline
KCGen-KT & \textbf{0.812} & \textbf{0.723} & \textbf{0.724} & \textbf{0.569}\\
w/o Correct Sol. & 0.789 & 0.674 & 0.704 & 0.529 \\
w/ Incorrect Sol. & 0.773 & 0.651 & 0.700 &  0.516\\
w/o KC Loss & 0.791 & 0.680 & 0.709 & 0.540 \\
w/o ICL Ex. & 0.782 & 0.677 & 0.705 & 0.539 \\
Partial ICL Ex. & 0.810 & 0.723 & 0.717 & 0.562 \\
Code $\rightarrow$ AST & 0.784 & 0.691 & 0.715 & 0.546 \\ 
Generated Code& 0.807 & 0.706 & 0.721 & 0.557 \\
\hline
\end{tabular*}
}
\vspace{-0.3cm}
\caption{Ablation study on CodeWorkout.}
\label{tab:ablation}
\end{table}

\noindent \textbf{Ablation Study} We conduct a comprehensive ablation study of all components of KCGen-KT on CodeWorkout, which we choose as the representative dataset due to its more balanced score distribution and cleaner, method-only code format. For ablation settings requiring code submissions for KC generation, we report results using two code solutions since LLM-generated code tends to follow similar problem-solving strategies; we adopt the most fine-grained KC abstraction and manually verify the correctness of all generated solutions. We first examine the importance of different inputs for KC generation. Including correct submissions is crucial; removing them results in a performance decrease on both KT tasks, indicating that the problem statement alone is insufficient to capture all necessary KCs. Similarly, in-context examples play a critical role—removing them leads to overly abstract KCs that are less effective for KT. Interestingly, replacing the complete problem–KC mapping in the in-context examples with only two randomly selected KCs yields comparable performance, suggesting that minimal human involvement is sufficient.

We further analyze alternative design choices for KC generation. Incorporating both correct and incorrect solutions during KC generation in order to capture mistake-related KCs. However, this ablation hurts performance, likely because KCs derived from incorrect solutions encode errors or misconceptions, thereby introducing noise into the soft-token-based model. Replacing actual student submissions with LLM-generated submissions also results in a minor performance decrease, which suggests that actual student code is highly diverse and captures a more complete set of skills required in each problem. Removing the KC loss term similarly results in lower performance on both tasks, indicating that regularization on KC mastery values is essential for accurate student modeling during training. We also see that switching submissions to the abstract syntax tree (AST) representation as input to the LLM decreases performance, where we see the generated KCs tend to be less detailed. This result can be explained by LLMs not being heavily pre-trained on AST representations of code. We also investigate the impact of the number of student submissions on KT performance (see Appendix~\ref{sec:ablation}).

\begin{figure*}[t]
  \centering
  \begin{subfigure}[b]{0.31\textwidth}
    \includegraphics[width=\linewidth]{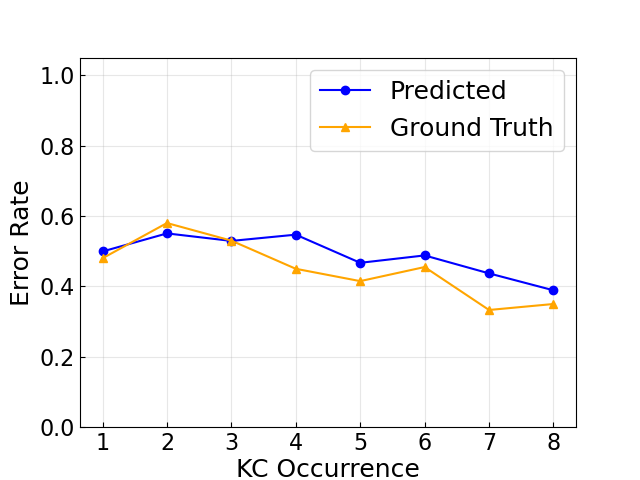}
  \end{subfigure}
  \hfill
  \begin{subfigure}[b]{0.31\textwidth}
    \includegraphics[width=\linewidth]{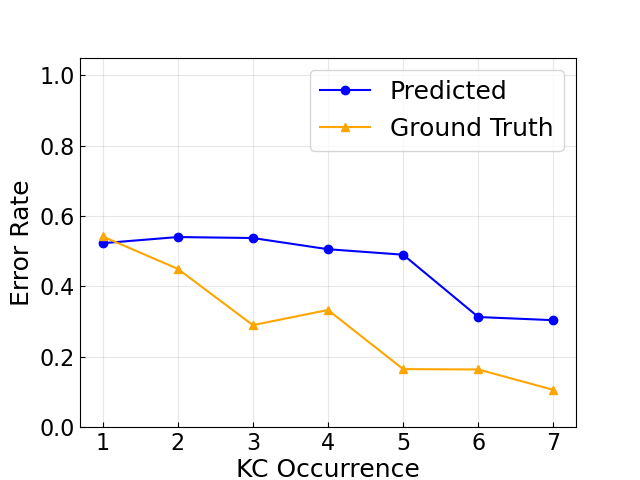}
  \end{subfigure}
  \hfill
  \begin{subfigure}[b]{0.31\textwidth}
    \includegraphics[width=\linewidth]{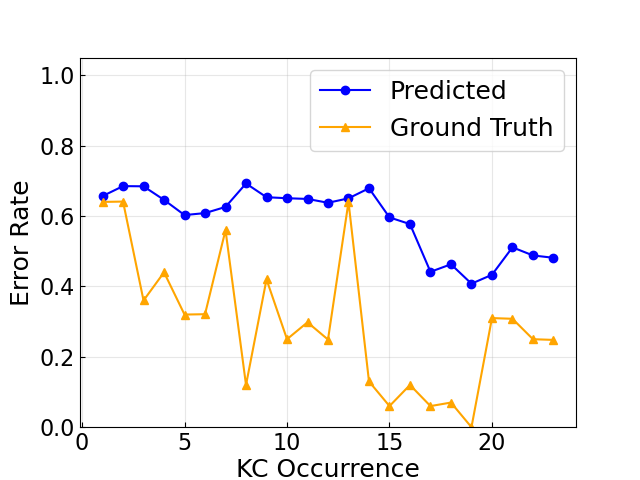}
  \end{subfigure}
  \vspace{-.1cm}
\caption{Representative learning curves for three generated KCs (Equality Comparison, String Length Determination, and For Loop Iteration), showing a generally decreasing error rate over attempts. Our KCs result in better model fit (0.21 vs.\ 0.18 in $R^2$) than human-written KCs under cognitive models \cite{Pavlik_Koedinger}.}
\label{fig:three_table}
\end{figure*}

\subsection{Qualitative Evaluation}
\noindent \textbf{Learning Curve Analysis} A common method to assess the quality of KCs is examining the alignment with the \emph{power law of practice}, which states that the error rate should decrease as the amount of practice on certain KCs increases~\cite{newell2013mechanisms,snoddy1926learning}. Hence, we compare the error rate across attempts at KCs with the estimated student KC mastery levels from KCGen-KT on CodeWorkout. We prompt GPT-4o to label KC correctness on incorrect student submissions. We prompt GPT-4o to (1) reason about the errors, (2) fix the errors, and (3) assign a binary correctness label to each KC, indicating whether the student made an error on this KC (See Appendix \ref{sec: prompt} for the exact prompt). In our experiments, we found that empirically, these KC-level error labels are mostly accurate; however, a formal human evaluation is necessary, which we defer to future work. 
%
To plot the curves, at attempt $t$, we average the binary correctness label over all students on the problem that represents their $t$-th attempt at the KC. We calculate the predicted error rate similarly, using KCGen-KT to estimate the mastery level of each student on each KC at each time step and taking the complement. 

Figure~\ref{fig:three_table} shows three representative learning curves among all generated KCs. In all cases, both ground truth and predicted error rates generally decline as attempts increase, aligning with the power law of practice. The first predicted learning curve closely aligns with the ground truth, demonstrating KCGen-KT's ability to accurately capture student learning progressions. The second predicted learning curve matches the ground truth error rates in trend, but has higher overall values. The third curve further exacerbates this discrepancy, for a KC that appears more frequently in the dataset. The ground truth error rates decrease overall but have significant fluctuation, making it difficult to fit the predicted learning curve. The likely reason is the non-linear problem sequencing in CodeWorkout; as students skip questions or attempt them in different orders, a single attempt index captures varying difficulty levels, making the estimation noisy.

For a more quantitative evaluation, following prior work \cite{Pavlik_Koedinger}, we fit PFA models on each KC. LLM-generated KCs achieve a weighted $R^2$ of $0.21$ versus $0.18$ using baseline KCs on CodeWorkout, which shows that LLM-generated KCs better fit the power law of practice. \\

\noindent \textbf{Case Study} To demonstrate the interpretability of our model, we provide a qualitative case study in Appendix \ref{sec: case_study}. Table~\ref{tab:qualitative_case_study} illustrates how predicted KC mastery levels correlate with specific errors in generated student code. This example further highlights that student errors in their responses to open-ended problems can be diverse and can be partly explained by their KC mastery deficiencies. \\

\begin{figure}[h]
  \centering
  \includegraphics[width=\linewidth]{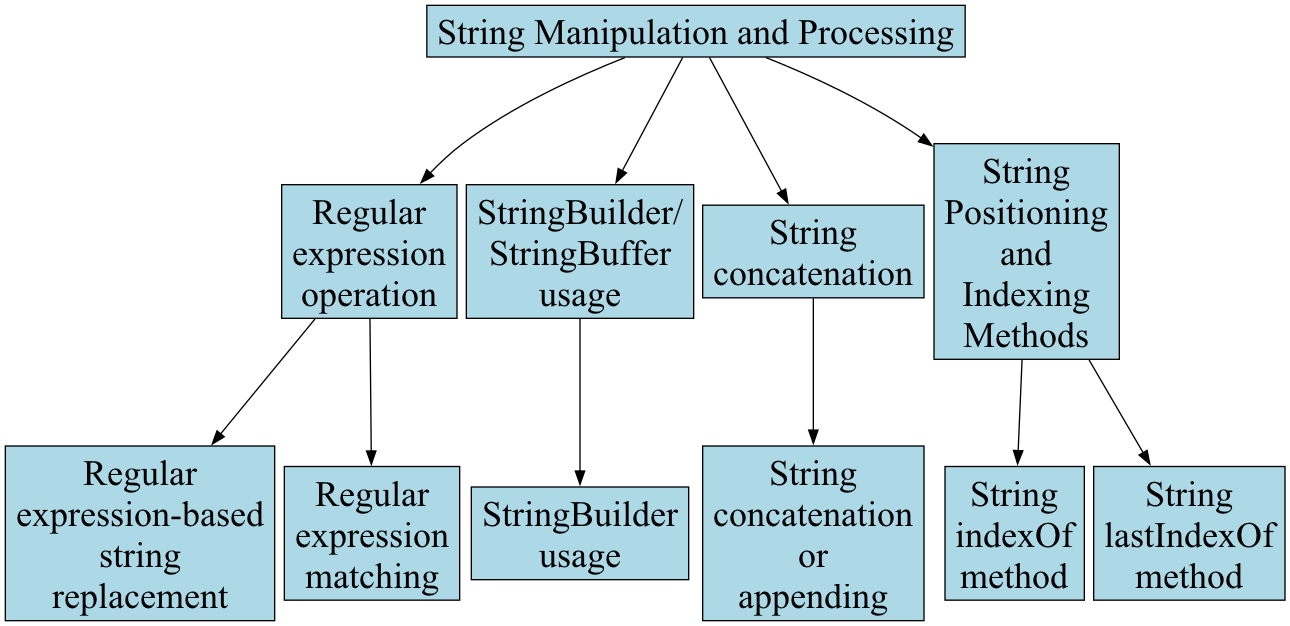}
  \caption{A section of the generated KC ontology (related to \texttt{Strings}), at different abstraction levels.}
  \label{fig:ont_tree}
\end{figure}

\noindent \textbf{KC Ontologies} 
We show a portion of the CodeWorkout KC ontology subtree obtained from KCGen-KT, for string-related concepts, in Figure~\ref{fig:ont_tree}. The root node shows the KC at the highest abstraction level; going down the tree from there, we see how the KCs identified increase in granularity at each level. To build the ontology tree, we start from the top-level KC and identify all KC labels from the next level that are semantically included in it, and do it iteratively for all KCs. This mapping defines the parent-child relationships between KCs across different abstraction levels. 

This example demonstrates the controllable abstraction property of our KC generation pipeline, where adjusting the number of clusters directly controls the granularity of the generated KCs.

\section{Human Evaluation} 
We perform a human evaluation to assess the quality of the generated KCs and the accuracy of problem-KC mappings. We recruit two annotators with experience in teaching college-level programming in Java. Since their expertise lies in Java-based instruction, we focus the evaluation on all $50$ CodeWorkout problems and provide them three separate tasks, each targeting a distinct aspect. 

To evaluate the \textbf{interpretability} of KCs, we provide the generated KCs for each problem and ask annotators to indicate whether each KC is clearly understandable from an instructor’s perspective. On average, $94.6\%$ of baseline KCs and $98.6\%$ of LLM-generated KCs are labeled as interpretable. Inter-annotator agreement is adequate: for example, the two human annotators reach a Cohen's Kappa score of 0.594 when labeling LLM-generated KCs, indicating (borderline) moderate-to-good agreement. This result suggests that the quality of generated KCs is comparable to or more informative than the human-created baseline KCs, highlighting their potential for instructional use in practice.

To evaluate problem-KC mappings, we evaluate \textbf{precision} and \textbf{recall}, measuring whether labeled KCs accurately and completely capture the skills required by each problem. Since there is no ``ground-truth'', for precision, we ask annotators to label each KC as relevant or irrelevant to the problem. For recall, we ask them to perform a head-to-head comparison between two sets of KCs and rate, in binary format, whether the LLM-generated set has less coverage or equal-or-greater coverage of essential KCs relative to the baseline. We then compute the average proportion of relevant KCs per problem and the proportion of cases where the generated set achieves equal or greater coverage.

Results show that the average precision is $92.5\%$ for the baseline set and is $93.2\%$ for the LLM-generated set with an inter-annotator agreement of $89.3\%$. For recall, the LLM-generated set is judged to have equal or higher coverage in $96\%$ of the cases, with $92\%$ agreement. We note that surpassing the baseline human-authored KC set on these tasks is highly challenging, making these results particularly encouraging. Overall, these findings indicate that LLM-based KCs are both interpretable and comparable to human-created baselines. See Appendix \ref{sec: human_eval} for the detailed annotation rubrics. 

\section{Conclusions and Future Work} 
In this paper, we presented an automated, LLM-based pipeline for KC generation and tagging for open-ended programming problems. We also developed an LLM-based KT framework, KCGen-KT, which leverages the textual content of KC descriptions. Through extensive experiments on two real-world coding datasets, we showed that KCGen-KT outperforms human-written KC labels on KC generation and existing state-of-the-art KT methods on predicting future student performance. We also show that LLM-generated KCs lead to better learning curves than human-written ones. A human evaluation shows that the generated KCs and problem-KC mappings are reasonably accurate for programming instructors. 

There are many avenues for future work. First, a human-in-the-loop KC generation method may address the limitation that KCGen-KT sometimes misses KCs shared across problems due to generating KCs independently for each problem. Second, we can explore integrating the learning curve fit directly into the objective to explicitly optimize for the power law of practice. Third, we plan to investigate applying our approach to other student modeling domains such as dialogues~\cite{scarlatos2024exploringknowledgetracingtutorstudent}, mathematics~\cite{ozyurt2024automated}, and science~\cite{moore2024automated}.

\section*{Acknowledgments}
This work is partially supported by the NSF under grants 2237676, 2418655, 2418657, and 2418658.

\newpage 

\section*{Limitations}
We note several technical and practical limitations. First, our automatic KC generation pipeline relies on in-context examples, requiring at least one human-written example to produce low-level KCs; in the absence of such examples, zero-shot prompting tends to yield more general KCs, as LLMs are not explicitly trained for this task. Second, because problems may involve multiple KCs, obtaining reliable ground-truth KC labels remains challenging, and our current labeling process 
would benefit from further validation. Third, our evaluation is limited to computer science education; extending the KC generation pipeline and KCGen-KT model to other domains, such as mathematics, would help assess generalizability. Finally, while we conduct human evaluation, the ultimate value of KC labeling lies in improving student learning outcomes, motivating future classroom studies comparing LLM-generated and human-authored KCs in downstream personalization settings.

\section*{Ethical Considerations}
There are several potential societal benefits to our work. Primarily, accurate automated KC generation can greatly benefit educational assessment: it can reduce the manual burden on educators in topic selection and KC design, thereby allowing more time and resources to be devoted to personalized student support, and it can enable more accurate student modeling with interpretability and focus on particular KCs. There are several potential risks to our work as well. As with many AI-assisted educational technologies, there is a concern that such systems could be perceived as replacing human educator jobs. We emphasize that our approach is intended to support, rather than replace, educators. A more immediate risk concerns the quality of automatically generated KCs. If the generated KCs are inaccurate, overly abstract, or misaligned with instructional goals, they may negatively affect student learning by misrepresenting required skills. To mitigate this risk, we recommend that automatically generated KCs be reviewed by domain experts before deployment in real educational settings.


\bibliography{custom}

\appendix
\section{Related Work}
\label{sec:rw}
\subsection{Knowledge Component Generation} 
Traditional methods for KC creation and tagging rely on human domain experts to identify the knowledge requirements for solving a problem~\cite{bier2014approach}, a highly time-consuming process. Recent work has proposed automated approaches for KC discovery and tagging, employing data-driven approaches including the Q-matrix method~\cite{barnes2005q}. 
In programming, \cite{hosseini2013javaparser} uses a rule-based parser to obtain ASTs with KCs identified at their lowest ontological level, ~\cite{rivers2016learning} defines KCs as nodes in an AST followed by a learning curve analysis to identify KCs that students struggle with in Python programming, \cite{hoq2024towards} uses an AST-based neural network to identify student misconceptions, \cite{shi2024knowledge} presents a deep learning approach for KC attribution, and \cite{shi2023kc,shi2024knowledge} learn latent KCs, lacking textual descriptions, by training deep learning models on KT data enforced with priors from pedagogical theory.
Recent advances in LLMs have inspired automated approaches for descriptive KC generation in dialogues~\cite{scarlatos2024exploringknowledgetracingtutorstudent}, math~\cite{ozyurt2024automated}, science~\cite{moore2024automated}, and learning curve studies~\cite{10.1145/3754508.3754514}. However, we're among the first approaches to present an automated LLM-based pipeline for KC generation and tagging for open-ended programming problems.

\begin{table*}[t]
\centering
\renewcommand{\arraystretch}{1.1}
\scalebox{0.92}{
\begin{tabular}{p{0.55\linewidth}|p{0.45\linewidth}}
\toprule
\multicolumn{2}{p{1.05\linewidth}}{Problem: Return true if the given string contains ``bob'', but where the middle `o' character can be any character.}\\
\midrule
Example Solution 1 & Example Solution 2 \\
\midrule
\begin{lstlisting}
public boolean bobThere(String str) {
    for (int i = 0; i < str.length() - 2; i++) {
        if (str.charAt(i) == 'b' && str.charAt(i + 2) == 'b') {
            return true;
        }
    }
    return false;
}
\end{lstlisting} &
\begin{lstlisting}
public boolean bobThere(String str) {
    boolean hasBobPattern = 
    str.contains("b") && str.matches(".*b.b.*");
    return hasBobPattern;
}
\end{lstlisting} \\
\bottomrule
\end{tabular}}
\vspace{-.3cm}
\caption{Example programming problem from the CodeWorkout dataset with two valid solution strategies. Although both solutions correctly solve the problem, they rely on different KCs.}
\label{tab:mot_example}
\end{table*}

\subsection{Knowledge Tracing}
KT has been extensively studied within the student modeling literature~\cite{kt}.
The classic KT task aims to estimate a student's mastery of KCs from their responses to past problems and use these estimates to predict their future performance.
Classic Bayesian knowledge tracing methods~\cite{ktcomparepardos,yudelson} use latent binary-valued variables to represent student KC mastery.
With the widespread adoption of neural networks, multiple deep learning-based KT methods were proposed with limited interpretability since student knowledge is modeled as hidden states in these networks.
Most of these methods use long short-term memory networks~\cite{lstm} or variants~\cite{dkt,saint+, 10.1145/3589334.3645582}, with other variants coupling them with memory augmentation~\cite{dkvmn}, graph neural networks~\cite{gikt}, or attention networks~\cite{akt,sakt, 10.1145/3589334.3645373}.
KT methods have been applied to many different educational domains, including programming~\cite{hoq2023sann,codedkt,pkt}. Recent work has attempted to leverage LLMs to develop generative KT methods predicting exact student responses to programming problems~\cite{tiktoc,infooirt,okt}. ~\cite{article} uses LLM to automatically construct Q-matrices capturing fine-grained KC relationship in KT, while their work lacks comparison with code-specific KT baselines, and does not explore LLM-based KC generation. To the best of our knowledge, we are the first to present an LLM-based KT method for programming problems that leverages the textual content of KC descriptions, modeling interpretable student mastery levels on each KC, for improved KT performance.

\section{Motivating Example}
\label{example: motivation}

Table~\ref{tab:mot_example} shows an example problem in the CodeWorkout dataset that highlights diversity in student-written code submissions. Although both code submissions are correct, they rely on different valid strategies leveraging different KCs: Solution 1 involves for loops and logical comparisons, whereas Solution 2 relies on regular expression matching. This example illustrates why KC identification and labeling can be quite challenging for open-ended problems compared to multiple-choice questions, where automated KC identification via LLMs has been studied in recent work \cite{moore2024automated,ozyurt2024automated}.

\section{Experimental Setup}
\label{sec: experiment}
As detailed in \ref{2.2}, we use a LLM as our backbone and we use the instruction-tuned version of Llama $3$ with $8$B paramters as our model. We use the Parameter Efficient Fine-Tuning (PEFT) library from HuggingFace~\cite{wolf-etal-2020-transformers} to load Llama 3 and fine-tune it via Low-Rank Adaptation (LoRA)~\cite{hu2022lora} ($\alpha=256$, $\text{rank}=128$, $\text{dropout}=0.05$) using 8-bit quantization. We use the AdamW \cite{loshchilov2018decoupled} optimizer for LLM and $W_m$ parameter with a batch size of $32$ and the RMSprop optimizer for the LSTM, and perform a grid search to determine the optimal learning rate. In KCGen-KT, we set different learning rates for different model components: $1e-5$ for Llama 3, $5e-4$ for the LSTM model, and $1e-4$ for the $W_m$ and $b_m$ parameters.  KCGen-KT converges within $12$ training epochs on both datasets, with each epoch taking $80$ minutes on the CodeWorkout dataset and $300$ minutes on the larger FalconCode dataset when trained on an NVIDIA L40S 48GB GPU. 

For the baseline model, TIKTOC*, which also uses an LLM as the backbone of the model, we use the same setup for a fair comparison: we use the same Llama model with PEFT to load it and fine-tune it with LoRA ($\alpha=256$, $\text{rank}=128$, $\text{dropout}=0.05$) using 8-bit quantization. The optimizers and learning rate used are the same as KCGen-KT, except the learning rate for the LSTM model is $1e-4$. For all models that perform the code prediction task, we use greedy decoding to generate student codes. 

For metrics, we use the rou\_auc\_score and f1\_score from scikit-learn library to compute AUC and F1 metrics, respectively. For codeBLEU we adopt the official implementation provided in the microsoft/CodeXGLUE. In addition, we use scipy library to perform the hierarchical agglomerative clustering. All software we use in the development of this work is open source. We are consistent with the terms and intended use of all software and with OpenAI API.

\section{Ablation Study}
\label{sec:ablation}
We explore the impact of the number of student submissions provided to the LLM on the quality of generated KCs and KT performance on CodeWorkout. For this ablation, we use the most fine-grained abstraction level. Table~\ref{tab:ablation_no_sol} shows the results, where we see that using fewer submissions, such as one and two, results in worse performance, while the performance increases and saturates after more than five submissions. This result can be explained by a smaller number of student submissions failing to capture diverse solution strategies for some problems, thus resulting in an incomplete KC set. As the number of submissions increase, the set of initial KCs before clustering more or less stays the same, and the performance mostly depends on the abstraction level instead.

\begin{table}[h]
\centering
\vspace{-0.1cm}
\small
\begin{tabular*}{\linewidth}{c @{\extracolsep{\fill}} cccc}
\hline
No. of Sol. & AUC & F1 & Acc & CodeBLEU\\
\hline
1 & 0.804 & 0.713 & 0.705 & 0.563\\
2 & 0.812 & 0.723 & 0.724 & 0.569 \\
5 & \underline{0.815} & 0.726 & \textbf{0.737} & \underline{0.572} \\ 
7 & 0.812 & \underline{0.726} & \underline{0.727} & \textbf{0.573} \\ 
10 & \textbf{0.816} & \textbf{0.727} & 0.725 & 0.566 \\ 
\hline
\end{tabular*}
\vspace{-0.3cm}
\caption{Ablation study of KCGen-KT on CodeWorkout on different number of solution codes provided during KC Generation. 
Best performance in \textbf{bold} and second best is \underline{underlined}.
}
\label{tab:ablation_no_sol}
\end{table}

\begin{table*}[ht]
\centering
\renewcommand{\arraystretch}{0.96}
\vspace{-.1cm}
\scalebox{.97}{
\begin{tabular}{p{0.45\linewidth}|p{0.35\linewidth}|p{0.1\linewidth}}
\toprule
\multicolumn{3}{p{1\textwidth}}{Problem: Given an array of ints, return true if the array contains two 7s next to each other, or there are two 7s separated by one element, such as with $\{7, 1, 7\}$. Otherwise, return false.}\\
\midrule
\multicolumn{1}{c}{Predicted Student Code Submission} & \multicolumn{1}{c}{LLM-generated KC} & \multicolumn{1}{c}{Mastery}\\
\midrule
\multirow{-1.8}{*}{\begin{lstlisting}[style=customJava]
public boolean has77(int[] nums){
    for (int i = 0; @i < nums.length - 1;@ i++){
        if (nums[i] == 7 && nums[i + 1] == 7){
            return true;
        }
        else if (nums[i] == 7 && @nums[i + 2] == 7@){
            return true;
        }
    }
    return false;
}
\end{lstlisting}
} & {For loop iteration} & \multicolumn{1}{c}{\textcolor{CornellRed}{26.1\%}} \\
& {Array indexing and assignment} & \multicolumn{1}{c}{\textcolor{CornellRed}{28.1\%}} \\
& {Boolean logic} & \multicolumn{1}{c}{\textcolor{OliveGreen}{51.9\%}} \\
& {Logical operators} & \multicolumn{1}{c}{\textcolor{OliveGreen}{57.6\%}} \\
& {Numerical comparisons} & \multicolumn{1}{c}{\textcolor{OliveGreen}{59.9\%}} \\
& {If and else if statement} & \multicolumn{1}{c}{\textcolor{OliveGreen}{74.4\%}} \\
\bottomrule
\end{tabular}}
\vspace{-0.3cm}
\caption{Example showing low student knowledge on relevant KCs map to specific errors in predicted student code.}
\label{tab:qualitative_case_study}
\end{table*}

\section{Case Study}
\label{sec: case_study}
Table~\ref{tab:qualitative_case_study} shows the estimated KC mastery levels and predicted code submission for a student on a problem from the CodeWorkout test set. The low student mastery level on KCs, ``For loop iteration'' and ``Array indexing and assignment'', results in a run-time error in the predicted code by indexing the array outside of its bounds. In contrast, the higher predicted mastery level of other KCs results in the correct implementation of the if and else if statements, proper use of boolean expressions, and accurate application of the logical AND operator. This example shows that informative KC descriptions generated by the LLM can help KCGen-KT make more accurate student code predictions. In practice, the predicted mastery level may offer instructors interpretable insights into the student’s understanding of specific programming concepts.

\section{Human Evaluation Details}
\label{sec: human_eval}
We conduct a human evaluation to assess (1) the interpretability of generated KCs as a measure of their instructional quality and (2) the precision and recall of the problem–KC mappings. Our evaluators were volunteers contacted through a research partner and were not compensated monetarily. They were made aware that their annotations would be used in scientific research in AI. The detailed annotation rubrics are provided below.

\subsection{KC Interpretability}
The instruction provided to annotators is as follows:

\noindent\fbox{%
    \parbox{\dimexpr\linewidth-2\fboxsep-2\fboxrule\relax}{%
        \leftskip=1em\rightskip=1em
        For each KC, indicate whether it is clearly understandable from an instructor’s perspective to decide the KC’s interpretability using Yes or No.    
    }%
}

\subsection{Problem KC Mapping}
We consider the correctness of problem-KC mapping from both precision and recall perspective. Precision measures whether KCs assigned to a problem are relevant. Recall metric measures whether the KC set sufficiently covers the essential concepts required for solving the problem. Because no absolute ground-truth KC set exists for each problem, recall is measured using relative coverage between the baseline and LLM-generated KC sets. The instruction provided to evaluate the problem-KC mapping is as follow: \\

\noindent\fbox{%
    \parbox{\dimexpr\linewidth-2\fboxsep-2\fboxrule\relax}{%
        \leftskip=1em\rightskip=1em
        \textbf{Precision:} For each KC associated with a problem, indicate whether it is relevant to that problem in binary format (relevant/irrelevant).
        \par
        \textbf{Recall:} For each problem, compare the generated KC set and baseline KC set and assign a binary label indicating which set better captures the essential KCs. Label "Less coverage" if Set 1 covers fewer relevant KCs than Set 2 and label "Equal-or-greater coverage" if Set 1 covers more relevant KCs than Set 2. If both sets contain unique KCs that the other is missing which means neither list clearly dominates, select equal-or-greater coverage. 
    }%
}

\section{Prompt}
\label{sec: prompt}

\subsection{Prompt for KC Generation Pipeline}
We show the prompt used for the KC generation in Table \ref{tab:kc_generation_prompt} and the prompt used for cluster summarization in Table \ref{tab:kc_summarize_prompt}.

\subsection{Prompt for KC Correctness Labeling}
We show the prompt used for KC correctness labeling for the learning curve analysis in Table \ref{tab:kc_correctness_label}.

\subsection{Prompt for In-context Example Conversion}
We show the prompt to convert the baseline KC list into natural language terms in Table \ref{tab:kc_example}.

\begin{table*}[h]
\centering
\begin{tabular}{p{0.95\linewidth}}
\toprule
\textbf{System Message:} \\
You are an experienced computer science teacher and education expert. You are provided with a list of human-labeled knowledge components (KCs) associated with programming problems. Your task is to convert each KC into a equivalent natural language term. \\ \\

\textbf{User prompt:} \\
The KC list is: [If/Else, NestedIf, While,	For, NestedFor, Math+-*/, Math\%,	LogicAndNotOr,	LogicCompareNum,	LogicBoolean,	StringFormat,	StringConcat,	StringIndex,	StringLen,	StringEqual,	CharEqual,	ArrayIndex,	DefFunction]\\
\bottomrule
\end{tabular}
\caption{Example prompt for Baseline KC conversion}
\label{tab:kc_example}
\end{table*}


\begin{table*}[h]
\centering
\begin{tabular}{p{0.95\linewidth}}
\toprule
\textbf{System Message:} \\
You are an experienced computer science teacher and education expert. You are given a Java programming problem along with n sample solutions. Your task is to identify generalizable knowledge components (skills or concepts) necessary to solve such problems. \\ \\

A knowledge component (KC) is a single, reusable unit of programming understanding, such as a language construct, pattern, or skill, that contributes to solving a programming problem and can be learned or mastered independently.\\ \\

Please follow these steps: \\
1. Analyze each solution carefully, noting critical constructs. \\ 
2. Reflect step by step on how each solution maps to distinct programming KCs that are independent and reusable. \\
3. For each KC, generate a concise name and provide a one-sentence reasoning explaining why this KC is necessary based on the provided solutions. Use the provided examples as reference for the appropriate level of detail. Make sure KCs are generalizable and applicable to a wide range of similar programming problems without referencing problem-specific details. \\
4. Ensure each KC is atomic and not bundled with others. \\ \\

Your final response must strictly follow this JSON template: \\
$\{$
    "KC 1": {{"reasoning": "Reasoning for this KC (exactly 1 sentence)", "name": "Knowledge component name"}},
    "KC 2": {{"reasoning": "Reasoning for this KC (exactly 1 sentence)", "name": "Another specific knowledge component name"}},
    ...
$\}$\\ \\

\textbf{User prompt:} \\
Example 1: \\
Problem: Write a function in Java that implements the following logic: Given a string \texttt{str} and a non-empty word, return a version of the original string where all chars have been replaced by pluses (+), except for appearances of the word which are preserved unchanged.\\[0.5em]
Expected Output:
KC 1: If and else statement
KC 2: While loop
KC 3: Numerical comparisons
KC 4: String formatting
KC 5: String concatenation
KC 6: String indexing
KC 7: String length
KC 8: String equality comparison
 \\ \\
Now analyze the following problem using their solution code. \\
Problem: A sandwich is two pieces of bread with something in between. Write a Java method that takes in a string \texttt{str} and returns the string that is between the first and last appearance of \texttt{"bread"} in \texttt{str}. Return the empty string \texttt{""} if there are not two pieces of bread.\\[0.5em]

\textbf{First sample solution is:} \\
\begin{lstlisting}
public String getSandwich(String str){
    String bread = "bread";
    if (str.contains(bread) && str.length() >= 10){
        int first = str.indexOf(bread);
        int last = str.lastIndexOf(bread);
        String between = str.substring(first + 5, last);
        return between;
    }
    else{
        return "";
    }
}
\end{lstlisting} \\

Follow the instructions in system message. First, carefully examine the solutions and identify the important elements and patterns. Then, explicitly reason about what underlying knowledge components are required based on these solution codes. Finally, take the examples as reference and summarize your analysis clearly into generalizable and concise knowledge components. \\
\bottomrule
\end{tabular}
\caption{Example prompt for KC generation with one in-context example and one student solution}
\label{tab:kc_generation_prompt}
\end{table*}

\begin{table*}[ht]
\centering
\begin{tabular}{p{0.95\linewidth}}
\toprule
\textbf{System Message:} \\
You are an experienced computer science teacher and education expert. You will be provided with a list of knowledge components (KCs) that may vary in wording but sometimes refer to the same or related underlying concepts or skills. \\

The KCs will be given in the format: ["KC 1 name", "KC 2 name", ..., "KC k name"] \\ \\

Your task is to: \\
1. Carefully examine all the KCs in the list to ensure none are overlooked. \\
2. Reason explicitly whether the KCs collectively refer to the same underlying concept or skill, or if they are related but represent distinct or complementary aspects of a broader theme. \\
3. Based on your reasoning: \\
    - If the KCs refer to the same concept or skill, select one KC from the list that best represents the group — choose the one that is most clearly worded, generalizable, and inclusive of the others. \\
    - If the KCs are related but too distinct to be represented by a single KC, create a concise and meaningful summary name that captures the broader theme or category shared by the KCs. \\ \\

Return your output strictly in the following JSON format: \\
$\{$
  "reasoning": "...",        $//$ Exactly one sentence explaining your reasoning \\
  "representative kc": "..." , $//$ Selected KC if applicable, otherwise null \\
  "summary name": "..." ,       $//$ Summary name if representative KC not chosen, otherwise null
$\}$\\ \\

\textbf{User prompt:} \\
The knowledge components list is: [for loop iteration, while loop, array iteration] \\ \\

Now follow the instructions in system message and perform the task. \\
\bottomrule
\end{tabular}
\caption{Example prompt for Cluster summarization}
\label{tab:kc_summarize_prompt}
\end{table*}

\begin{table*}[ht]
\centering
\begin{tabular}{p{0.95\linewidth}}
\toprule
\textbf{System Message:} \\
You are an experienced computer science teacher and education expert. You are given a Java programming problem, an incorrect student submission, and a predefined list of general programming knowledge components (KCs) relevant to solving the problem. \\
The predefined KCs relevant to solving the problem will be given in a list using the format ["KC 1 name", "KC 2 name", ...]  \\ \\      

Your task is to: \\
1. Identify all key errors in the student's code, and describe each error in exactly one sentence. \\
2. Assess the student's mastery of each provided KC in the list based on the incorrect submission.\\
        - Reflect on the student's original incorrect code.  \\
        - For each KC, return a binary label which equals 1 if the student makes an error on this KC, and equals 0 if not. \\ \\

Your final response must strictly follow this JSON template: \\
$\{$
    "error reasoning": [
        "First error described in one sentence.",
        ...
    ],
    "KC error": {{
    "KC 1 name": 0/1,
    "KC 2 name": 0/1,
    ...
    }}
$\}$\\ \\

\textbf{User prompt:} \\
Problem: \\
Write a function in Java that implements the following logic: Given 2 ints, a and b, return their sum. However, sums in the range 10..19 inclusive, are forbidden, so in that case just return 20. \\
Incorrect submission: \\
\begin{lstlisting}
public int sortaSum(int a, int b){
    if (a + b <= 10 && a + b >= 20)
        return 20;
    else
        return a + b;
}
\end{lstlisting} \\ \\

The knowledge components are: [Basic arithmetic operations, Logical operators, If and else if statement, Numerical comparisons]\\

Follow the instructions in system message to evaluate the KCs. \\

\bottomrule
\end{tabular}
\caption{Example prompt for KC correctness labeling}
\label{tab:kc_correctness_label}
\end{table*}

\end{document}